\documentclass[runningheads]{llncs}

\usepackage{eccv}

\usepackage{eccvabbrv}

\usepackage{graphicx}
\usepackage{caption}
\usepackage{booktabs}

\usepackage[accsupp]{axessibility}  
\usepackage{multicol}
\usepackage{multirow}

\usepackage{hyperref}

\usepackage{orcidlink}

\begin{document}

\title{Temporal Residual Jacobians for\\ Rig-free Motion Transfer} 


\author{Sanjeev Muralikrishnan\textsuperscript{1}, Niladri Dutt\textsuperscript{1}, Siddhartha Chaudhuri\textsuperscript{2}, Noam Aigerman\textsuperscript{3}, Vladimir Kim\textsuperscript{2}, Matthew Fisher\textsuperscript{2}, Niloy  Mitra\textsuperscript{1,2}}
\author{Sanjeev Muralikrishnan\inst{1}\orcidlink{0000-0002-3556-5007} \and
Niladri Dutt\inst{1}\orcidlink{0000-0002-7423-2221} \and Siddhartha Chaudhuri\inst{2}\orcidlink{0009-0009-8588-1436} \and Noam Aigerman\inst{3}\orcidlink{0000-0002-9116-4662 } \and Vladimir Kim\inst{2}\orcidlink{0000-0002-3996-6588} \and Matthew Fisher\inst{2}\orcidlink{0000-0002-8908-3417} \and Niloy J. Mitra\inst{1,2\orcidlink{0000-0002-2597-0914}}}

\authorrunning{Muralikrishnan et al.}

\institute{\textsuperscript{1}University College London, \textsuperscript{2}Adobe Research, \textsuperscript{3}University Of Montreal}

\maketitle

\newcommand{\name}{Temporal Residual Jacobians\xspace}
\newcommand{\shape}{$X$\xspace}
\newcommand{\initshape}{$X_0$\xspace}
\newcommand{\shapet}{$X_t$\xspace}
\newcommand{\step}{$t$\xspace}
\newcommand{\sig}{\ensuremath{m_z}\xspace}
\newcommand{\motionT}{\ensuremath{M_t}\xspace}
\newcommand{\JposeT}{\ensuremath{J^P_t}\xspace}
\newcommand{\Jposefirst}{\ensuremath{J^P_0}\xspace}
\newcommand{\features}{\ensuremath{C}\xspace}
\newcommand{\Jresidualfirst}{\ensuremath{J^R_0}\xspace}
\newcommand{\JresidualT}{\ensuremath{J^R_t}\xspace}
\newcommand{\Jt}{\ensuremath{J_t}\xspace}
\newcommand{\funcPose}{\ensuremath{f_P\xspace}}
\newcommand{\funcResidual}{\ensuremath{f_R\xspace}}
\newcommand{\attnPoseW}{\ensuremath{E^P_W\xspace}}
\newcommand{\attnResidualPast}{\ensuremath{E^R_{W-1}\xspace}}
\newcommand{\JposeW}{\ensuremath{J^P_W}\xspace}
\newcommand{\JresidualWpast}{\ensuremath{J^R_{W-1}}\xspace}
\newcommand{\timeW}{\ensuremath{T_W}\xspace}
\newcommand{\timeWpast}{\ensuremath{T_{W-1}}\xspace}

\newcommand{\vnudge}{\vspace*{-.15in}}

\newcommand{\shapesig}{\ensuremath{\beta}\xspace}
\newcommand{\shapesigPred}{\ensuremath{\beta'}\xspace}
\newcommand{\betaloss}{\ensuremath{L_{\beta}}\xspace}
\newcommand{\common}{\ensuremath{G}\xspace}
\newcommand{\centroid}{\ensuremath{\mathbf{c}}\xspace}
\newcommand{\stateForA}{\ensuremath{e_{acc}}\xspace}
\newcommand{\stateForV}{\ensuremath{e_{vel}}\xspace}
\newcommand{\stateA}{\ensuremath{s_{i_{acc}}}\xspace}
\newcommand{\stateV}{\ensuremath{s_{i_{vel}}}\xspace}
\newcommand{\jointsT}{\ensuremath{R_t}\xspace}
\newcommand{\funcAcc}{\ensuremath{f_\text{acc}^J}\xspace}
\newcommand{\loss}{\ensuremath{\mathcal{L}}\xspace}

\newcommand{\vode}{VertexODE\xspace}
\newcommand{\njf}{NJF(\ensuremath{M_t})\xspace}
\newcommand{\softSmpl}{SoftSMPL\xspace}

\newcommand{\smk}[1]{\textcolor{blue}{#1}}
\newcommand{\todo}[1]{{\color{red}[TODO: #1]}}
\begin{abstract}
We introduce Temporal Residual Jacobians as a novel representation to enable data-driven motion transfer. 
Our approach does not assume access to any rigging or intermediate shape keyframes, produces geometrically and temporally consistent motions, and can be used to transfer long motion sequences. Central to our approach are two coupled neural networks that individually predict local geometric and temporal changes that are subsequently integrated, spatially and temporally, to produce the final animated meshes. The two networks are jointly trained, complement each other in producing spatial and temporal signals, and are supervised directly with 3D positional information. During inference, in the absence of keyframes, our method essentially solves a motion extrapolation problem.
We test our setup on diverse meshes (synthetic and scanned shapes) to 
demonstrate its superiority in generating realistic and natural-looking animations on unseen body shapes against SoTA alternatives. Supplemental video and code are available at \url{https://temporaljacobians.github.io/}. 
\end{abstract}
\section{Introduction}
\label{sec:intro}

A major challenge in character animation is to transfer the motion of a source (skeletal) system to a diverse range of target characters in a realistic manner. The traditional approach to achieve this involves using a 	{\em rig}, which connects a skeleton to the character’s surface and manages the surface motion through a variety of constraints and parameters. Target movement is then conveyed from the skeleton to the surface by either simulating the physics of the muscles and fat or by using tailored sets of skinning weights. These weights can be manually created by skilled artists or derived from pre-existing rigged models~\cite{AnimSkelVolNet,RigNet:20}. 
Despite its simplicity, rigging can be time consuming to set up and complicated to transfer to new target shapes; it may also fail to accurately capture dynamics.

There has been growing interest in developing surface deformation techniques that are more flexible and efficient than traditional rigging-based approaches and can utilize available volumes of full-body motion capture data. One possibility is to train data-driven methods to learn a low-dimensional parameterization, such as a morphable humanoid template~\cite{SMPL:2015} or a neural space deformation~\cite{Yifan:NeuralCage:2020}, to provide controllable handles for shape and pose-aware manipulations. However, such methods do not capture continuity over time and can overlook subtle motion dynamics essential for enhancing the realism of the generated motion sequences. To add time-dependent effects, corrective vertex deformations, similar to DMPLs~\cite{SMPL:2015} and SoftSMPL~\cite{santesteban2020softsmpl}, have been introduced in multistage workflows. Unfortunately, such methods do not account for elementwise temporal inter-relations 
and have limited generalization to unseen characters. 

\begin{figure}[t!]
  \includegraphics[width=\textwidth]{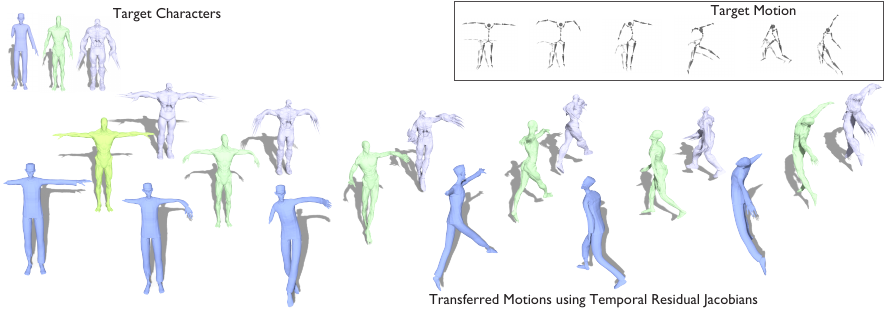}
  \caption{
  Given a stick figure dance motion (top-right), \textit{Temporal Residual Jacobians} retarget the animation to unseen, unrigged meshes~(top-left) across time, producing realistic motion dynamics. Please refer to the supplemental webpage for videos. Our method can be trained on limited data, does not require rigged models or skinning weights during training or inference, and does not assume paired sequences or registration to any canonical template mesh. The method was trained on other bodyshapes: no target characters were seen during training. All results in the paper and supplemental material were obtained with automatic feature correspondences and without any postprocessing or smoothing applied. 
  }
  \label{fig:teaser}
  \vnudge 
\end{figure}

We aim to transfer a source motion, expressed as joint angles on a stick figure, onto a target shape, specified as a (rig-free) mesh. We want to do so without access to any rigging on the target shape, and we also want to avoid any fixed template or morphable shapes. 
A desirable solution should address several challenges: (i)~handle rig-free meshes and/or scans with arbitrary topology; (ii)~produce plausible transfers to diverse shapes; (iii)~achieve continuity over space and time and thus avoid artifacts (e.g., broken meshes, jittery motion); and (iv)~work with long motion sequences without significant drift. 
The first two problems are partially handled by rig-free pose transfer (i.e., transferring a pose to a target mesh) methods~\cite{aigerman2022neural,liao2022pose,wang2023zeroshot}. 
While such methods produce plausible single frames, they lead to jittery motion transfer and motion-induced geometric artifacts like shearing.

We propose a novel approach that learns local spatio-temporal changes to produce natural-looking motion transfers. Technically, we achieve this via a new representation in the form of \textit{Temporal Residual Jacobians} that temporally links spatial predictions and is directly supervised using example motion sequences. We jointly train two neural networks to individually predict local spatial and temporal changes. They are coupled by spatial integration with a differentiable Poisson solve, and temporal integration with a neural ODE. A {\bf key technical insight} is that instead of having the neural ODE predict per-frame mesh deformations, it is more effective to predict initial deformations independently of time via a base posing model (we use Neural Jacobian Fields~\cite{aigerman2022neural}), and then have the neural ODE predict {\em residual deformations}, linked over time, as corrective factors that improve temporal coherence. 
Figure~\ref{fig:teaser} shows how a stick-figure control motion produces target motions for different characters, without requiring registration to any standard template or character rigs.

We evaluate the effectiveness of our method for motion generation to different character bodies (e.g., humanoids, animals, Mixamo characters, scans) and different motions (walk, run, jump, punch, dance). We compare our approach to alternatives, when available. 
We provide qualitative and quantitative results using the AMASS~\cite{AMASS:ICCV:2019}, COP3D~\cite{sinha2023common}, and 4DComplete~\cite{li20214dcomplete} datasets. 

In summary, our primary contributions are: (i)~a novel method that enables motion transfer via Temporal Residual Jacobians and can be trained directly using positional data; (ii)~local predictors that can be integrated in space and time to create natural looking character animations; and (iii)~a robust pathway to transfer realistic character motion without the need for explicit rigging or learning a parameterization using any canonical template shape.

\section{Related Work}

\paragraph{Parametric Shape Deformation.} These methods express 2D or 3D shapes as a known function of a set of common parameters, and model deformations as variations of these parameters. Such methods include cages, explicit~\cite{ju2005meanvalue} or neural~\cite{Yifan:NeuralCage:2020}, blendshapes~\cite{lewis2014blendshape}, skinned skeletons~\cite{jacobson2014skinning}, Laplacian eigenfunctions~\cite{rong2008spectral}, and several other variations. Linking the parameters to the shape's surface often requires manual annotation of weights (commonly known as weight painting) in 3D authoring tools. Alternately, given sufficient data (i.e., meshes, rigs, skinning weights), end-to-end training can produce realistic neural rigs, as demonstrated by Pinocchio~\cite{baran2007automatic}, RigNet framework~\cite{RigNet:20}, skinning-based human motion retargeting~\cite{marsot2023correspondencefree}, and skeletal articulations with neural blend shapes ~\cite{li2021learning}. Unsupervised shape and pose disentanglement ~\cite{zhou2020unsupervised} proposes to learn a disentangled latent representation for shape and pose, which can be further used to transfer motion using shape codes. This requires meshes to be registered and have the same connectivity. 
 To animate these parametrized shapes through time, the parameters are varied over time and the dynamic weights animate the mesh. These methods require access to body templates and/or rigs and can produce results that are jittery due to loose coupling of the individual frame predictions. 
 
\vspace{-2mm}
\paragraph{Dynamic Motion.}
It is possible to model temporal surface effects by simulating the underlying soft tissues using finite element methods (FEM) \cite{chadwick1989layered,fan2014active}. This direct simulation is typically slow and requires artists to design the underlying bone and muscle structure~\cite{abdrashitov2021musculoskeletal}. Approaches have been developed to overcome the stiffness problem in FEM to accelerate simulating these systems~\cite{modi2020efficient} or to solve the problem in a lower-dimensional subspace~\cite{park2008data}. For the specific case of rigged human characters, Santesteban et al.~\cite{santesteban2020softsmpl} add soft-tissue deformation as an additive per-vertex bump map on top of a primary motion model; AMASS~\cite{AMASS:ICCV:2019} imparts secondary motion using the blending coefficients of the DMPL shape space~\cite{SMPL:2015}; and Dyna~\cite{pons2015dyna} learns a data-driven model of soft-tissue deformations using a linear PCA subspace. These efforts, however, assume access to primary motion via a skeleton rig and are restricted to humans, registered to a canonical template. Complementary dynamics~\cite{zhang2020cd,benchekroun2023fast} models physically-based secondary motion in the subspace complementary to that spanned by an animation rig. This approach adds automatic secondary motion to arbitrary animated objects, but requires the target shape to be rigged, is not designed for deformation transfer (the base animation is part of the input), and is tied to a specific hand-coded secondary physics model. DeepEmulator~\cite{zheng2021deep} achieves a similar effect using a local-patch-based neural network to learn the secondary behavior, but again requires the primary motion as input and does not support deformation transfer.

\vspace{-2mm}
\paragraph{Discrete Time Motion Models.}
Given their ability to model time, deep recurrent neural networks have also been used to model shape sequences. Fragkiadaki et al.~\cite{fragkiadaki2015erd} use LSTMs to predict short human joint motions given initial frames. Harvey et al.~\cite{harvey2020robust} leverage LSTMs for in-betweening to predict intermediate joint motions. He et al.~\cite{he2022nemf} learn a motion field for joints through time. In all these methods, the mesh itself is deformed via rigging, and since joint motion does not encode bodyshape, they do not sufficiently represent secondary motion. Also, being discrete time representations, these approaches must train on large datasets of joint motion. Qiao et al.~\cite{qiao2018learning} instead use mesh convolutions with LSTMs to deform vertices through time. In our work, we use neural ODEs as they can model time continuously instead of discretizing time and modeling the sequence using LSTMs. Further, vertex-based deformation models are susceptible to artifacts like normal-inversion as we demonstrate in our evaluation section.

\section{Approach}
Given an unrigged, triangulated mesh of a 3D character, we aim to animate it by motion transfer from an available motion described by relative joint angles (stick figure motion) at each time step. The relative joint angles are represented as Euler angles and are defined at each joint with respect to its hierarchy in SMPL's kinematic tree (cf.~~\cite{he2022nemf}). We obtain these joint-based motion representations from the AMASS dataset \cite{AMASS:ICCV:2019}. Since we aim to animate the mesh itself from the joints' motion, we seek to learn a mapping from the joint representation to the positions of the given mesh's vertices, and to do so at each time step while ensuring we generate a smooth and artifact-free mesh animation. We supervise our setup with ground-truth meshes from the AMASS dataset \cite{AMASS:ICCV:2019}.

We parameterize such a character \shape $\in \mathbb{R}^{N \times 3}$ as assigning positions to each of its $N$ vertices of the underlying mesh. Thus to impart a motion to a mesh, we perform this assignment at every time step. Given a shape \initshape as a triangulated mesh, in its initial pose configuration, along with per-frame pose configuration \motionT that describes the target pose at time $t$, we aim to predict the shape \shapet at each time \step generating the full motion sequence; essentially learning a mapping from relative joint orientations to mesh deformations (typically, from ~30-50 joints to 50-100k vertices). We aim for a data-driven method that generalizes to new characters and does \textit{not} rely on shape rigging or intermediate shape keyframes. Another desirable property is to do so with limited data (e.g., working with 5-10 motion examples), as obtaining full-body 3D motion data is non-trivial. 

Our key observation is that we can robustly train neural networks to predict changes \textit{local} in both space and time, which can then be integrated across space, using a differentiable Poisson solve; and across time, using Euler equations to handle an ODE numerically.
Integrating across space helps maintain plausibility of the entire shape, while integrating across time helps model a realistic, time-coherent motion. Further, local predictions help higher generalization capability to unseen body shape; while the spatio-temporal integration ensures smoothness of the sequence, without jitters or undesirable shearing artifacts. 
Our end-to-end differentiable formulation allows training the networks directly using example training sequences, without requiring factorized spatial and temporal motion signals. 
For spatial handling, we extend the representation from local deformation encoding~\cite{defTransfer:sigg:04} and affine mapping framework Neural Jacobian Fields (NJF) \cite{aigerman2022neural}, which learns an affine transformation field that is sampled at each mesh face and integrated into vertex positions via a Poisson solve. For temporal coherence and consistency in these predicted Jacobians, we couple temporal signals across the character motions using a neural ODE framework~\cite{neuralODE:18} via a novel Temporal Residual Jacobian representation. Predicting {\em residual deformations} that correct the predictions of a base model for time-coherence turns out to be more effective than trying to make the base model itself time-coherent.

\subsection{Overview}
\paragraph{Training:} Since our goal is to map temporally varying joint angles to full mesh animation, during training we assume this mapping is available i.e., we have one-to-one mapping across time between joint orientations and the corresponding meshes. Thus, we use the motion sequences in AMASS \cite{AMASS:ICCV:2019} as our training dataset for humanoid shapes. Thus, during training, our algorithm takes in the joint angles at each time step, the mesh itself at time $t=0$ in its starting pose, and the full mesh sequence is used for supervising our neural networks.

\paragraph{Inference:} At inference time, our algorithm simply takes in an unrigged character provided as a triangulated mesh. Our trained method is then used to animate this mesh from joint-sequences of motions sampled from AMASS. These meshes can be obtained in-the-wild or from character datasets like Mixamo. For shapes in the AMASS dataset since we have the ground-truth motions, we evaluate our framework quantiatively. We only show qualitative motion generation for other shapes in the work, as there is no ground-truth solution.

\paragraph{Motion and Shape parameters:} The motion sequences in AMASS were obtained from live captures of subjects performing different motions; these were then parametrized in SMPL's \cite{SMPL:2015} shape and pose space, thus providing a mapping from joint orientations to 3D mesh pose - i.e., vertices of the mesh oriented to match the pose represented by the provided joint orientations. We use this mapping to supervise our framework. Specifically, this data framework allows us to sample motions across both shapes and poses, by varying the respective low-dimensional parameters $\beta$ and $\alpha$. In our work, the $\alpha$ translate to joint angles at each time step and we use the source live-capture subject's $\beta$ as is, and pass it as a conditioning variable to our method.

\subsection{Preliminaries}
Our framework uses two components -- one to learn spatial mesh deformations (re-posing) and the other to learn temporal signals to generate a temporally coherent mesh motion. We describe these below. 

\paragraph{Neural Jacobian fields,} as introduced in \cite{aigerman2022neural}, encode local deformations by defining a field via map $\phi$, defined over space, that can be sampled at the $N$ vertices of the surface mesh. Specifically, given $\phi$, we compute Jacobians in the basis of the triangles of the surface mesh as,
\begin{equation} \label{eq:jacobian}
    J_i = \phi \nabla_i^T
\end{equation}
where $\nabla_i$ is defined as the gradient of triangle $t_i$ in its basis $B_i$ defined at the triangle's centroid. Thus, given a learned map, we obtain each triangle's (estimated) affine transformations $J_i$. Recovering the vertex positions from this per-triangle assignment of affine transformations is then done via a least-squares formulation that reduces to solving a Poisson system given by (cf.~\cite{poissonMesh:04}),
\begin{equation} \label{eq:poisson}
    \phi^* = L^{-1} \rho \nabla^T J
\end{equation}
where $\rho$ is the mesh's mass matrix, $L=\nabla^T \rho \nabla$ is the cotangent Laplacian, and $J$ is a stack of (estimated) Jacobians $J_i$. 
This solution gives a unique mesh up to a translation, which we fix using \initshape, the mesh at $t=0$. Mesh positions directly supervise the neural map $\phi$ via a differentiable Poisson solver~\cite{Nicolet2021Large}. 

Thus, NJF allows us to learn affine deformations of a given mesh. In our work, the map $\phi$ is a trainable neural network that predicts local deformations of the mesh's triangles. Given conditioning parameters, the trained map can then be used to predict the Jacobians of each triangle in the mesh at each time step, generating a time-coherent animated 3D mesh.

\begin{figure*}[t!]
    \centering    \includegraphics[width=.95\linewidth]{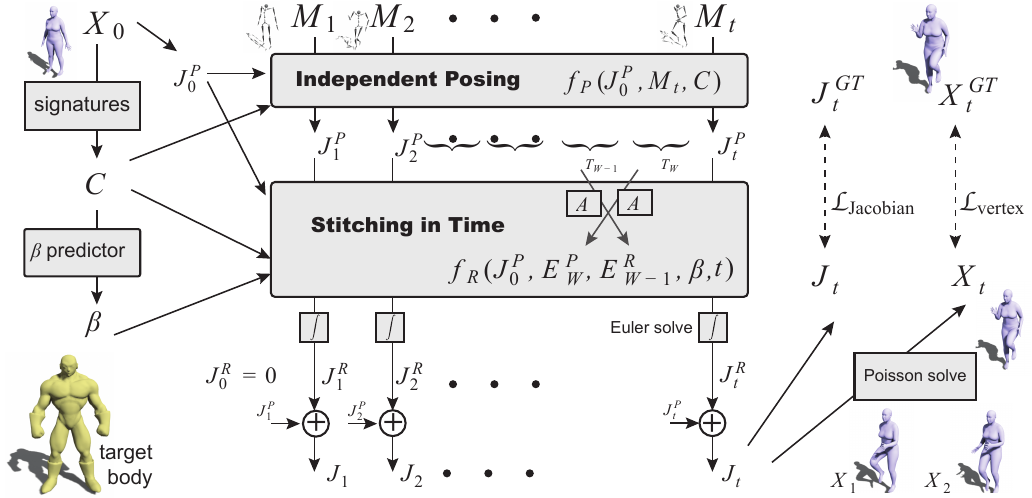}
    \caption{
    {\bf Method overview. }
    Starting from input stick figure motion ($\{M_i\}$) and a target body shape ($X_0$), Residual Temporal Jacobians makes local predictions, using primary $f_P$ and residual $f_R$ MLPs, to predict spatial and temporal changes to per-triangle Jacobians.
     These are then integrated in space, via a Poisson solve, and in time, via numerical Euler stepping, to predict motion dynamics at time frame $t$. These two learnable modules are trained simultaneously with only direct object-level supervision using a combination of positional and Jacobian losses. We use the ground-truth meshes from AMASS to supervise the predictions.
The time $t$ input is positionally encoded.}
    \label{fig:pipeline}
    \vnudge 
\end{figure*}

\paragraph{Neural ODEs} Chen et al.~\cite{neuralODE:18} represent differential equations with neural networks to model the dynamics of time-varying systems (motion sequences in our setting). Specifically, given a function $f(t;\theta) := \frac{\partial J}{\partial t}$, where $f$ is a neural network with parameters $\theta$ and $t$ being time, the variable $J$ can be integrated out at $t$ as
\begin{equation}\label{eq:diffeq}
    J(t) = \int_0^t \frac{\partial J}{\partial t} dt = \int_0^t f(t;\theta) dt + J_0.
\end{equation}

The neural network can be directly supervised with $J$ values as the network output utilizes a black box differential equation solver without explicitly discretizing the dynamic system. This leads to better estimation, benefits from constant memory updates (as opposed to explicit hidden states in RNNs, LSTMs, etc.), and trades numerical precision for speed.

Our work seeks to learn how the triangle Jacobians move through time. Thus, our neural ODE operates in the Jacobian space, with Jacobians being the variables to integrate across time. 

\subsection{Motion Transfer with Space-time Integration}
Without keyframes, we have to solve a temporal extrapolation problem. Our goal is to produce shape-preserving realistic motion sequences, as well as to learn such a motion model from a very sparse dataset. Not surprisingly, if each frame were independently predicted, the resulting motion would not be consistent across time, resulting in a jittery and artifact-ridden sequence (see Section~\ref{sec:results}). We, therefore, propose a novel framework wherein independent frame predictions are improved by providing temporal signals from a neural ODE. The two networks work in lockstep, each boosting the other's predictions, and are trained together. 

\paragraph{Independent posing:} We estimate each frame's Jacobians independent of temporal information. Given a shape \initshape, pose configuration (relative joint angles with respect to SMPL's kinematic tree) \motionT and features describing each mesh face, we predict \JposeT as the primary Jacobians at time $t$. Specifically, 
\begin{eqnarray}
    \Delta \JposeT &=& \funcPose(\Jposefirst, \motionT, \features; \theta_{P}) \\
    \JposeT &=& \Jposefirst + \Delta \JposeT
    \label{eq:posefunc}
\end{eqnarray} 
where \Jposefirst is the first frame Jacobian computed from $X_0$, \motionT is the pose configuration given as relative joint angles, and \features are per-face features defined at the centroid of the mesh faces,  of the mesh at $t=0$. In our tests, as features \features, we jointly learned PointNet features on the centroids and normals of \initshape and augmented them with pre-computed Wave Kernel Signatures~\cite{aubry2011wave}. Our feature network (a shallow PointNet) learns geometric features that enable mapping to an unseen shape via correspondence in this learned feature space. We note that conditioning the posing network on \Jposefirst and adding the predicted delta via a residual connection significantly improves pose prediction and generalization to unseen pose configurations. We additionally note that the Jacobians are computed in the local basis defined at each face's centroid. Thus, \Jposefirst is the Jacobian of the identity deformation in the local basis (a rotation). Henceforth, we do not update the basis in the sequence and express all Jacobians on this chosen basis.

In practice, we analytically compute the triangle Jacobians of only the first frame \initshape in the sequence. We then predict the Jacobians at each time instance $t$ in the coordinate frame of \initshape and solve Equation~\ref{eq:poisson} to obtain the shape \shapet at time $t$. Importantly, we augment NJF with temporal learning signals, by linking these independent per-frame predictions via a neural ODE, as described next.

\paragraph{Stitching across time:} Our key observation is that learning \textit{local} changes in time generalizes better to unseen shapes. Central to our method is a neural ODE that provides temporal training signals to the primary NJF and integrates across time to learn a smooth, arbitrary-length motion sequence. We found independent per-frame predictions are prone to artifacts and do not generalize well (see Section~\ref{sec:results}). 
We also observe that a neural ODE, in isolation, cannot predict the entire sequence due to the drift problem inherent to estimating functions using numerical ODE methods. Specifically, given an initial state and per-frame control parameters (in our case, joint angles), predicting the mesh sequence is an extrapolation problem. As such, ODEs are prone to drifting away from the underlying function. This problem is exacerbated as the length of the motion increases -- the longer the motion, the larger the accumulated drift.

As a solution, we propose a novel formulation to address both the incoherence of per-frame predictions and the drift problem. Specifically, we direct the neural ODE to learn only {\em Residual Jacobians} at each time step conditioned on the predictions from Eq~\ref{eq:posefunc} and on a window of past Jacobians. The Residual Jacobians are corrective factors which are directly added to the per-frame Jacobians. We predict Residual Jacobians local in time as outputs of a Neural ODE to ensure temporal coherence. This allows us to handle much longer motions, spanning 1-3k frames, without noticeable drift. We describe our method below.
\paragraph{ODE formulation:} To handle arbitrary length sequences, in the interest of memory and training speed, we train and infer a given sequence in windows of consecutive frames. A given sequence is broken into fixed window sizes. We initialize $J_0$ in Eq \ref{eq:diffeq} as the first frame's Residual Jacobian. Since the first frame is stationary and given as input, we set the first frame's Residual Jacobian as, 
\begin{equation}
        \Jresidualfirst = \textbf{0} \in \mathbb{R}^{3\times3}. 
\end{equation}
We then task the neural ODE to learn Residual Jacobian \JresidualT at each $t$, by extrapolating from \Jresidualfirst using Euler's integration. We then extract the final Jacobians in terms of the base Jacobians corrected by the Residual Jacobians as
\begin{equation}
        \Jt = \JposeT + \JresidualT.
        \label{eq:finaljacobians}
\end{equation}
We predict the residuals $\JresidualT$ by integrating over time the bodyshape-specific local changes predicted by $\funcResidual$ which is defined as,
\begin{equation}
    \frac{\partial \JresidualT}{\partial t} = \funcResidual(\Jposefirst,\attnPoseW,\attnResidualPast,\shapesig,t; \theta_R)
    \label{eq:funcResidual}
\end{equation}
where \shapesig is the shape signature that defines the given body shape, and \Jposefirst, as defined previously, are the Jacobians of the first frame; \attnPoseW and \attnResidualPast are attention encodings of current-window pose predictions and previous-window residual predictions. We integrate the local changes (see Eq~\ref{eq:funcResidual}) over time using Euler's method to obtain \JresidualT at each $t$ as,
\begin{equation}
    \JresidualT = \int_0^t \frac{\partial \JresidualT}{\partial t} dt + \Jresidualfirst 
=  \int_0^t \funcResidual(\Jposefirst,\attnPoseW,\attnResidualPast,\shapesig,t; \theta_R) dt.
    \label{eq:ode}
\end{equation}
Our jointly trained attention encoders are defined as,
\begin{eqnarray}
    \attnPoseW &=& A^P(\JposeW,\timeW) \\
    \attnResidualPast &=& A^R(\JresidualWpast,\timeWpast) 
\end{eqnarray}
where $A^P$ and $A^R$ are multi-head attention networks, \JposeW and \JresidualWpast are a block of sequential Jacobians in the current window $W$ and past window $W-1$, respectively; \timeW and \timeWpast are correspondingly blocks of time instances in these windows and are positionally encoded using time. In all our experiments we use a window size of 32 frames.

Note that we use the attention networks to encode a window of Jacobians to a single encoding. Since the encoding sizes are thus constant, we can handle arbitrary window/sequence lengths without overflowing memory. These encoders distill the current posed Jacobian predictions from Eq~\ref{eq:posefunc} and previously predicted Residual Jacobians obtained from Eq~\ref{eq:ode}. 

We pass the output of the attention networks as conditioning to Eq \ref{eq:funcResidual} to integrate and obtain the residuals in Eq \ref{eq:ode}. The predicted residuals are then added to the posed Jacobians in Eq \ref{eq:finaljacobians}. Finally, we spatially integrate the predicted Jacobians \Jt using a differentiable Poisson solve~\cite{Nicolet2021Large}, in the coordinate frame of the first frame, to obtain the predicted shape $X_t$ at $t$.

\paragraph{Loss function:} Our pipeline is trained end-to-end using only a shape loss over vertices of \shapet and a Jacobian loss. Thus, our final objective function is simply
\begin{equation}{\label{eq:lvertex}}
    \loss_{\text{vertex}} = \| X_t - X_{t}^{GT} \|^2 \quad \text{and} \quad 
   \loss_{\text{Jacobian}} = \| J_t - J_{t}^{GT} \|^2,
\end{equation}

\begin{equation}
    \loss = \loss_\text{vertex} + \alpha \loss_\text{Jacobian}.
\end{equation}
We use $\alpha=0.05$ in our tests.

\begin{figure*}[!t]
    \centering
    \includegraphics[width=\textwidth]{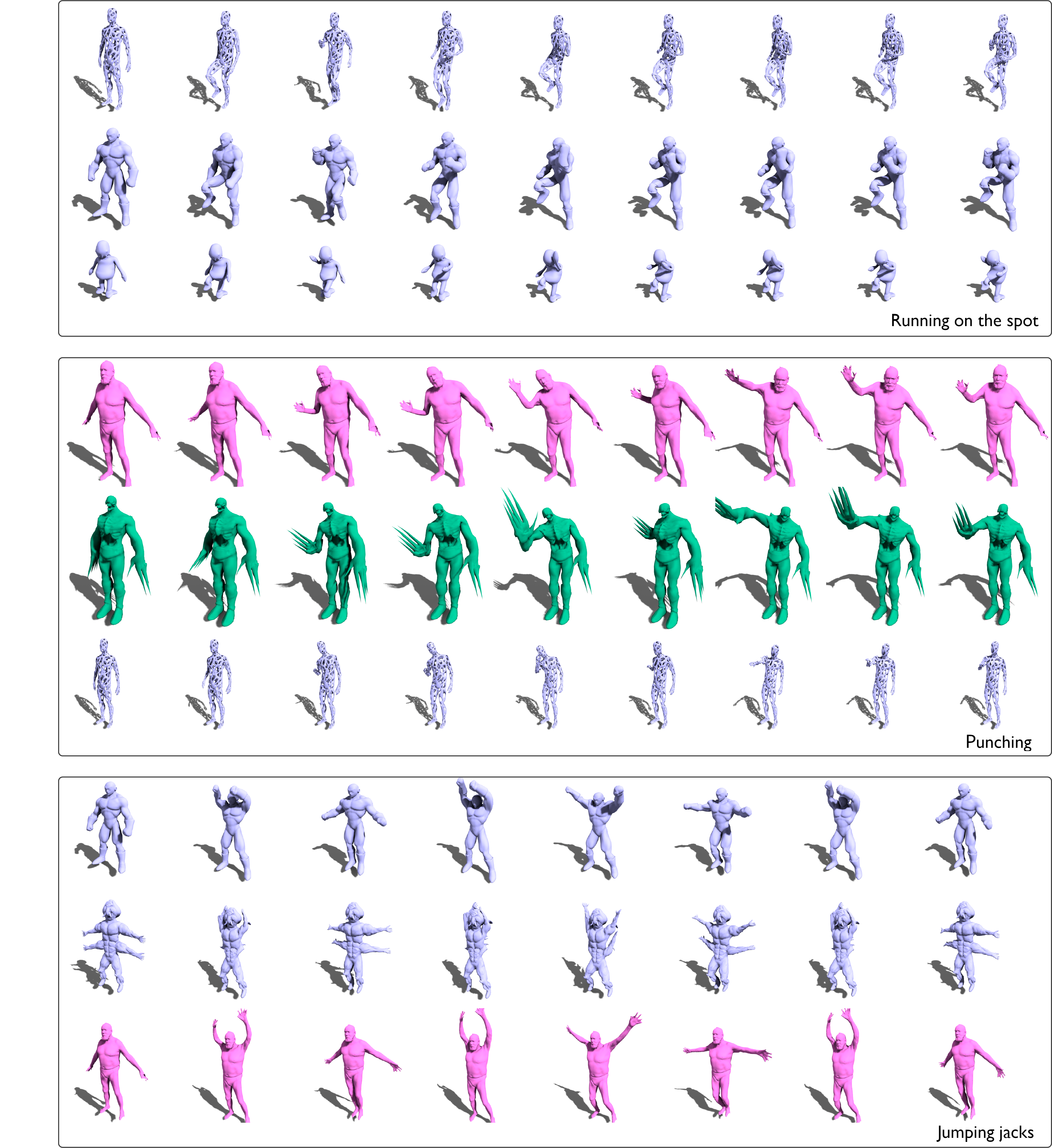}
    \caption{
    { \bf Generalization across bodyshapes.} We show results of different motion transfers on meshes found in-the-wild (blue), FAUST scans (pink) and Mixamo characters (green). We observe a smooth motion consistent with the target geometry in each case. Please see  supplemental materials.}
     \label{fig:joints_to_body}
     \vnudge
\end{figure*}

\begin{figure*}[b!]
    \centering
    \includegraphics[width=\textwidth]{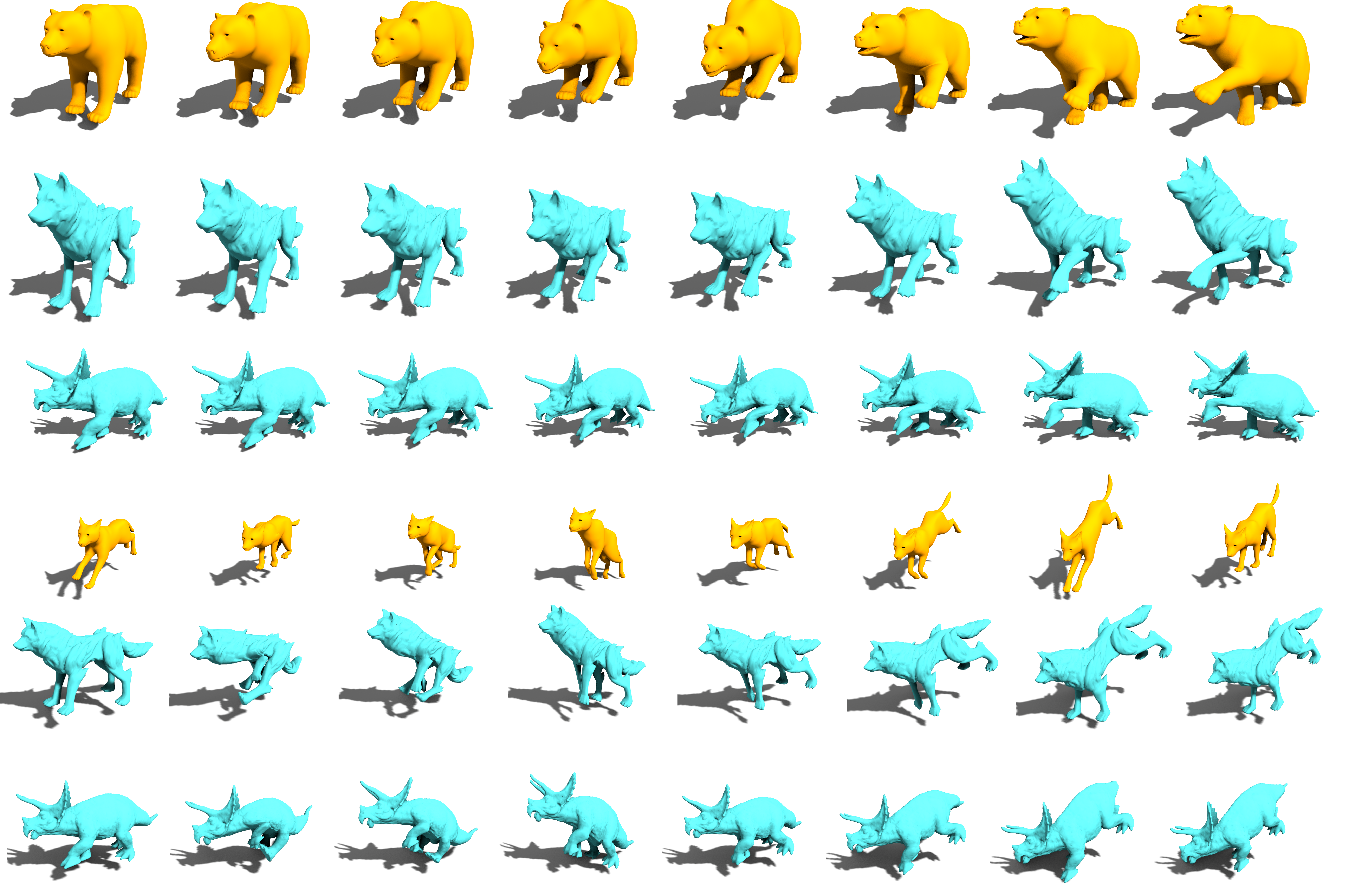}
    \caption{
    { \bf Generalization across shapes with very sparse training sets.}
     Here, we show motion transfer from two animal sequences (in yellow) sampled from the DeformingThings4D dataset~\cite{def4d}, to animal meshes found in the wild~(in blue). Our method was trained on only two sequences from this dataset and yet generates plausible motion transfer to unseen shapes. 
     Rigs were \textit{not} available to our algorithm at training and/or test time. 
     (Note: blue sequences have been slightly globally rotated for visibility.)}
     \label{fig:joints_to_body}
\end{figure*}

\section{Evaluation}
\label{sec:results}

\paragraph{Motion datasets.}
We train and evaluate our method on motion sequences from three different datasets. First, 
the AMASS dataset \cite{AMASS:ICCV:2019} for humanoid motions, which is based on the SMPL body model~\cite{SMPL:2015} that allows us to vary body-shapes by changing its shape parameter $\beta$, for a given sequence. Thus, we train our model for a given motion category (like running) to predict the sequences on varying body-shapes (indicated as the $\beta$-predictor in Figure~\ref{fig:pipeline}). During training, each sequence in the category is trained with only one body-shape. During inference, given an unknown test shape $X_0$ and a motion $\{M_t\}$, we synthesize the full body motion for the target shape. Second, we  also tested on a synthetic DeformingThings4D dataset~\cite{def4d}, which provides animal 4D meshes as deforming sequences. Finally, we also present results on motions extracted from video recordings of animals in the COP3D dataset~\cite{sinha2023common} where our inputs are meshes fitted to the video recordings to train \name.

\paragraph{Target shapes.} We evaluate our method on various forms of target shapes, namely sampled SMPL models, scans from the FAUST dataset, characters from the Mixamo library, various models from online 3D repositories (e.g., wolf, triceratop, monster, hole-man, etc.). Non-manifold inputs were converted to manifold meshes before running our algorithm. All correspondences were as inferred by the signature module (i.e., combination of PointNet features on centroids and normals of faces along with WaveKernel Signatures, previously defined as \features).

\paragraph{Implementation details.} All the networks we use are shallow MLPs to aid training speed. For our independent posing network, we use a 4-layer MLP with ReLU activation, with the final layer being linear. For the residual Jacobian prediction, we use a 3-layer MLP similarly with intermediate ReLU activations and a final linear layer. Both our attention networks follow the same architecture -- we use two attention heads, with a 32-neuron wide feed forward layer and 32-dimensional features for the keys and values. Our PointNet network from which we obtain per-face features similarly has 3 ReLU layers followed by a linear layer at the end. Our method and the baselines are trained until $\loss_{\text{vertex}}$ converges to less than 3e-4 or upto 300 iterations. On a 12 GB TitanX our typical training time is 6-8 hours, varying by the lengths of the sequences. 
\\
For the \texttt{walking} and \texttt{dancing} motions, we set the root orientation in AMASS to zero, so all subjects are front facing, making it easier to train (similar to \cite{he2022nemf}), as our Jacobian-based formulation cannot, on its own, infer global rotation and/or translation. We do not, however, predict global transformations; we obtain the global transforms from the source AMASS sequences and apply them to our final outputs at inference after appropriate scaling according to the target's height.

\paragraph{Qualitative results.} We show various examples of deformation transfer by our method in Figures~\ref{fig:teaser}, \ref{fig:joints_to_body}, and \ref{fig:results_plate}. Please refer to the videos in the supplemental webpage. Our method generalizes to new shapes of varied body types, including non-humans {\em completely unseen during training}, while preserving the source motion. We show examples on monsters, a 4-armed character, etc. Additionally, as different humans perform the same motion in different manners, we capture those varying dynamics in the deformation of the new shape as well. We can also learn and apply motions from animals, both from synthetically generated mesh sequences as well as motions extracted from monocular video recordings.

\begin{figure*}[h!]
\centering\includegraphics[width=.9\textwidth]{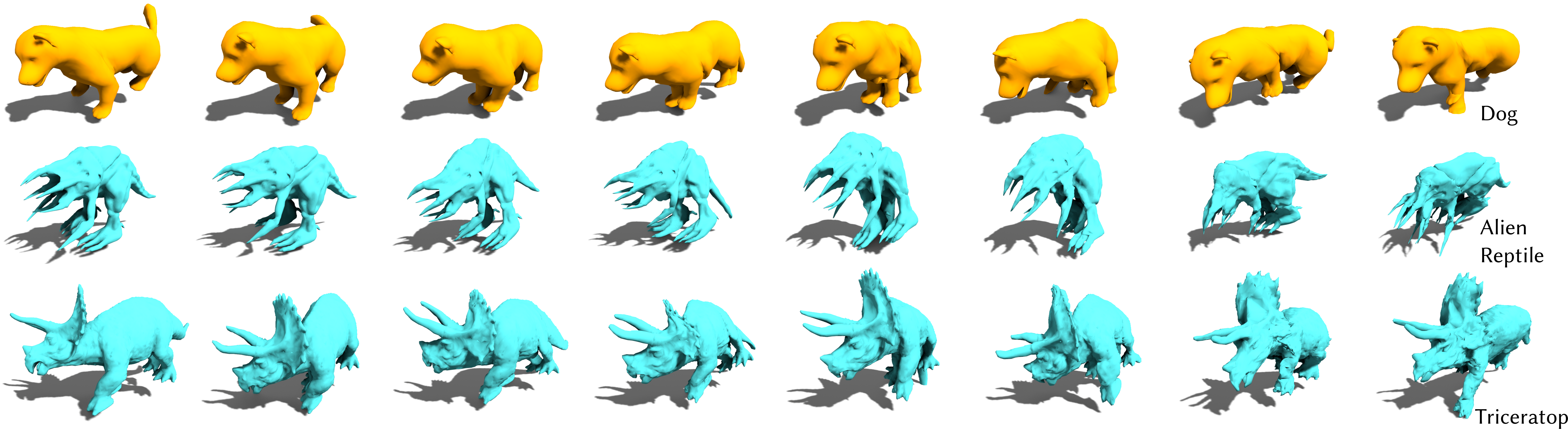}
    \caption{
    {\bf Motion transfer from COP3D dataset.} We train on only four sequences of dogs obtained from the COP3D dataset~\cite{sinha2023common}, which are monocular video recordings of animal motion, and transfer the observed regressed motion (in yellow) to creature meshes found in the wild (in blue).} 
    \label{fig:results_plate}
    \vnudge
\end{figure*}

\begin{figure*}[!h]
    \centering
    \includegraphics[width=.9\textwidth]{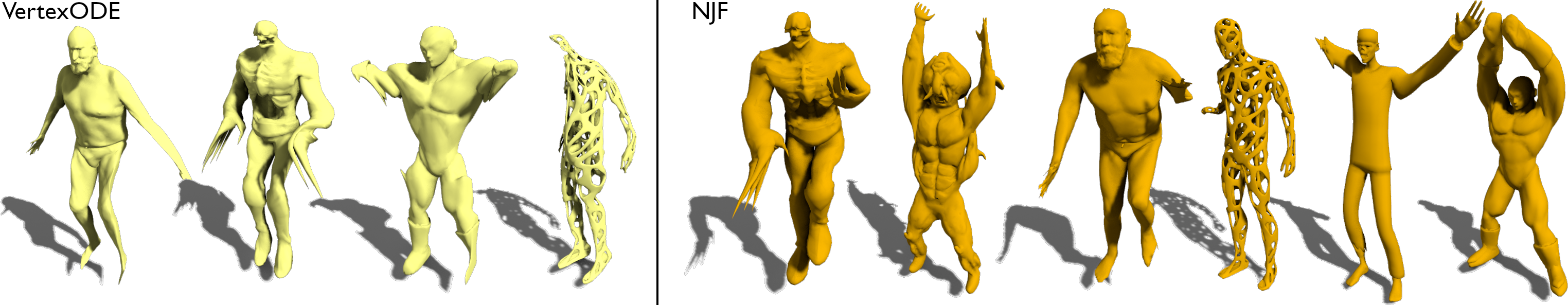}
    \caption{
    {\bf Observed artifacts in baselines.}
    For each motion, we show results from an intermediate frame of the motion transfer for our baselines. VertexODE (left, yellow) completely distorts the shape, while not following the target motion. NJF (right, brown) suffers from temporal discontinuity resulting in motion-driven geometric artifacts -- extended or shrunken parts as it tends to a linear path in later time steps -- and jitter.
    }
     \label{fig:result_comparison}
     \vnudge
\end{figure*}

\paragraph{Comparisons.} Our method \name, due to the space-time coupled formulation, is designed to produce natural-looking motion retargeting. Although 
we are unaware of any other method that performs the specific motion extrapolation problem (i.e., \textit{without} access to keyframes), we compare with possible baseline design variations. 
First,  \textit{\vode}, where an MLP predicts the rate of change of vertex positions (velocity) at time $t$, followed by a Neural ODE~\cite{neuralODE:18} that takes numerical steps by Euler's method to integrate to the vertex positions at $t$. This baseline directly displaces the vertices (i.e.,  without triangle Jacobians). As seen in Figure~\ref{fig:result_comparison}, this method produces significant artifacts and leads to degenerate shapes. It also loses the intended motion after some time, as seen in the videos. 
Second, \textit{\njf}, where we extend the original NJF framework to be additionally conditioned on per-frame relative joint orientations \motionT, to predict the Jacobians at each time step. Framewise prediction leads this approach to suffer from jitters (please refer to supplemental videos) and a tendency to overfit to training data. Although the individual frames are mostly plausible, they occasionally suffer from frame-level artifacts as shown in Figure~\ref{fig:result_comparison}. The primary limitation, however, is jittery output as seen in the video results since the frames are independently generated. Note that a method without spatial or temporal derivatives wherein a simple MLP is trained to predict vertex deformations given the initial geometry and time $t$ performs poorly, with numerous artefacts due to vertex-level predictions.

\paragraph{Quantitative comparison:}
We evaluate our method on several types of sequences and compare ours against other baselines in Table~\ref{table:error_compare} by computing the residual error with respect to the ground-truth sequences. We compare five different motion categories from the AMASS dataset - \texttt{running}, \texttt{jumping jacks}, \texttt{walking}, \texttt{hopping}, and \texttt{punching}. We achieve the best on all metrics.

\paragraph{Human motion to non-humanoid characters.}
We show results on human motion applied to non-humanoid characters (e.g., 4-armed monster, alien-reptile) in supplemental videos. Please note the quality of automatic transfer without manual landmark specification.

\begin{table}[t!]
\centering
\scriptsize
\caption{
    {\bf Quantitative evaluation.} 
    Average vertex-to-vertex error in cm, L2 error of predicted Jacobians and angular error of normals in degrees, measured against ground truth sequences, for different motion categories and averaged over multiple sequences within the same target motion category. 
    Here we compare against neural ODE~\cite{neuralODE:18} and an extended version of NJF~\cite{aigerman2022neural}. Lower values indicate better generalization.}
\label{table:error_compare}
\begin{tabular}{r|ccc|ccc|ccc|ccc|ccc}
\toprule
\multirow{2}{*}{\textbf{Method}} & \multicolumn{3}{c|}{\texttt{Jump}} & \multicolumn{3}{c|}{\texttt{Run}} & \multicolumn{3}{c|}{\texttt{Punch}} & \multicolumn{3}{c|}{\texttt{Walk}} & \multicolumn{3}{c}{\texttt{Dance}} \\
& L2-V & L2-J & L2-N & L2-V & L2-J & L2-N & L2-V & L2-J & L2-N & L2-V & L2-J & L2-N & L2-V & L2-J & L2-N \\
\midrule \midrule
\vode & 22.59 & 1.22 & 48.21 & 14.23 & 0.83 & 42.11 & 17.66 & 0.65 & 40.04  & 23.92 & 1.01 & 46.12 & 26.15 & 1.26 & 50.17 \\
\njf & \underline{5.52} & \underline{0.41} & {\underline 9.66} & \underline{3.61} & \underline{0.32} & \underline{7.38} & \underline{4.95} & \underline{0.38} & \underline{7.42} & \underline{6.85} & \underline{0.34} & \underline{8.12} &  \underline{4.86} & \underline{0.44} & \underline{11.24} \\
TRJ (Ours) & \bf{2.64} & {\bf 0.28} & {\bf 7.31} &  {\bf 1.73} & {\bf 0.24} & {\bf 5.63} & {\bf 2.86} & {\bf 0.26} & \bf{6.65} & {\bf 1.48} & {\bf 0.22} & {\bf 6.33} & {\bf 3.96 } & {\bf 0.37} & {\bf 9.88} \\
\bottomrule
\end{tabular}
\end{table}

\paragraph{Handling long sequences.} One advantage of our space-time coupled formulation is the ability to handle long motion sequences. We show examples of motion transfer from a few hundred to a couple of thousand frames (dance and walk) in the supplemental.

\paragraph{Design variation.} A seemingly possible variation of ours is to choose a residual Jacobian representation, but expressed in terms of the previous predicted frame. While this seems attractive, we found it very slow to train as we have to backpropagate through time, and convergence is very slow (when using moderate computational resources). Hence, we found this approach to be infeasible with the current memory requirements for attention (and transformer) modules.

\paragraph{Alternatives to Euler Solve.} We 
 experimented with higher-order ODE solvers, namely Runge-Kutta and Runge-Kutta-Fehlberg. However, we noticed no significant improvements, even with the noticeably increased training time. We use the simpler Euler's method for temporal integration for our tests. 
\section{Conclusion}

We have presented Temporal Residual Jacobians, a spatially-coupled, neural ODE-based, motion transfer framework conditioned on body type and target motion to produce local Jacobians that are subsequently integrated across space and time to deform the target geometry. The resultant motions are robust, realistic, and generalize to different body types. We extensively tested our method on both synthetic and real data captures, and enabled generic motion transfers to an extent which is not possible using existing methods.

Our method has limitations. 
(i)~We do not impose physics constraints, therefore, our animation can have self-intersections (see webpage videos). An interesting direction is to incorporate constraints for collision detection, e.g., via subspace-based contact handling~\cite{subspaceSelfContact:21}. We want to incorporate such an approach directly into the method, possibly via an attention mechanism, as motion dynamics are affected by earlier collisions. 
(ii)~Although our Temporal Residual Jacobians, along with windowed attention modules, keep drifts low, error still accumulates over long motion sequences. A possible solution is to couple our method with a keyframe-based workflow~\cite{WangEtAl:GarmentAuthoring:2019}.
(iii)~Finally, our current formulation implicitly establishes correspondences and uses them to infer temporal Jacobians for surface triangles. This step may incur errors as shown in project webpage (e.g., flapping ear of the mouse). These spurious correspondences can be overridden by artists, possibly directly by ``paintbrushing'' correspondences or materials (e.g.,  indicating that the ear of the bunny model should be floppy) or using semantic features learned from untextured meshes~\cite{texture23,dutt2023diffusion}. 

%
%
\noindent\textbf{Acknowledgement} This project has received funding from the UCL AI Center, gifts from Adobe Research, and EU Marie Skłodowska-Curie grant agreement No 956585.
\bibliographystyle{splncs04}
\bibliography{temporalResidualJac}
\end{document}